\pdfoutput=1

\documentclass[11pt]{article}

\usepackage{emnlp2021}

\usepackage{times}
\usepackage{latexsym}

\usepackage[T1]{fontenc}

\usepackage[utf8]{inputenc}

\usepackage{microtype}

\usepackage{amsmath,amsfonts,bm}

\def\eqref#1{equation~\ref{#1}}

\def\1{\bm{1}}

\DeclareMathAlphabet{\mathsfit}{\encodingdefault}{\sfdefault}{m}{sl}
\SetMathAlphabet{\mathsfit}{bold}{\encodingdefault}{\sfdefault}{bx}{n}

\usepackage{mathabx}
\usepackage{subfig}
\usepackage{float}
\usepackage{times}
\usepackage{soul}
\usepackage{url}
\usepackage{hyperref}
\usepackage[utf8]{inputenc}
\usepackage{caption}
\usepackage{graphicx}
\usepackage{multirow}
\usepackage{amsmath}
\usepackage{amsthm}
\usepackage{booktabs}
\usepackage{algorithm}
\usepackage{algorithmic}
\usepackage[normalem]{ulem}
\useunder{\uline}{\ul}{}
\title{An Empirical Study on Few-shot Knowledge Probing for\\ Pretrained Language Models}

\author{Tianxing He \\
  MIT \\
  \texttt{tianxing@mit.edu} \\\And
  Kyunghyun Cho \\
  New York University \\
  \texttt{kyunghyun.cho@nyu.edu} \\ \And  
  James Glass \\
  MIT \\
  \texttt{glass@mit.edu} \\
  }

\begin{document}
\maketitle
\begin{abstract}
Prompt-based knowledge probing for 1-hop relations has been used to measure how much world knowledge is stored in pretrained language models. Existing work uses considerable amounts of data to tune the prompts for better performance. In this work, we compare a variety of approaches under a few-shot knowledge probing setting, where only a small number (e.g., 10 or 20) of example triples are available. In addition, we create a new dataset named TREx-2p, which contains 2-hop relations. We report that few-shot examples can strongly boost the probing performance for both 1-hop and 2-hop relations.  In particular, we find that a simple-yet-effective approach of finetuning the bias vectors in the model outperforms existing prompt-engineering methods. Our dataset and code are available at \url{https://github.com/cloudygoose/fewshot_lama}.
\end{abstract}

\vspace{-0.2cm}
\section{Introduction}
\vspace{-0.15cm}

Large-scale unsupervised pretraining \cite{elmo18peters,jacob18bert,song2019mass,xlnet19zhilin,yinhan19roberta} of language models (LMs)  has been shown to greatly boost the performance of a wide variety of natural language processing (NLP) tasks. It is interesting to wonder how much world knowledge is embedded within these pretrained LMs. In the LAMA benchmark \citep{petroni2019knowlanguage}, templates (e.g., ``[X] was born in \texttt{<mask>} .'') are used to create natural-language prompts for 1-hop relations in an existing knowledge graph. The accuracy of whether the model can predict the right object token is treated as a proxy for how knowledgeable the pretrained LM is. This line of investigation \citep{poerner-etal-2020-e,kassner-schutze-2020-negated,kassner-etal-2021-multilingual,heinzerling-inui-2021-language} also points to an exciting potential application of using a pretrained LM as an implicit knowledge base.

\begin{figure}
    \centering
    \includegraphics[width=0.8\linewidth]{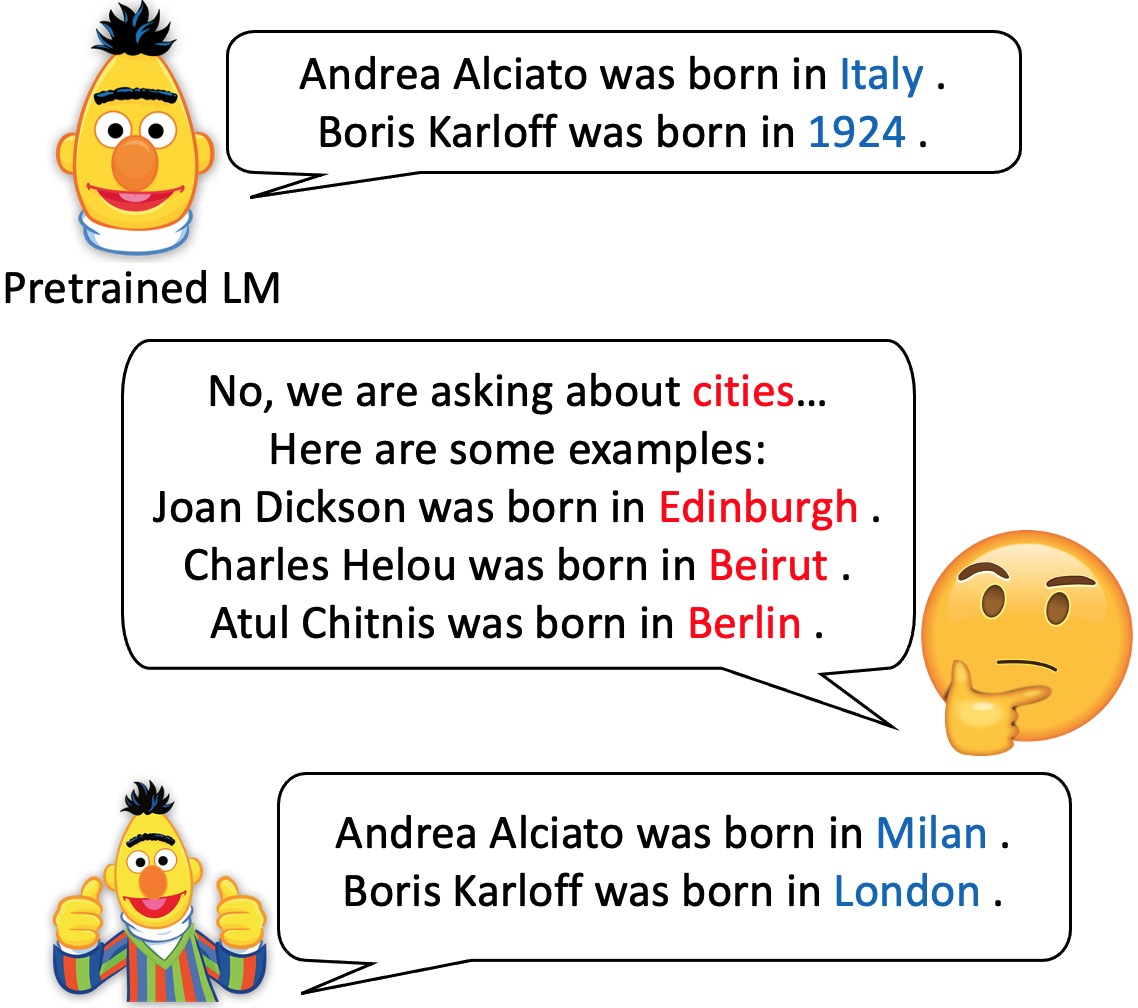}
    \vspace{-0.25cm}
    \caption{Few-shot examples can potentially correct the model's prediction in knowledge probing.}
    \label{fig:fewshot_motivation}
    \vspace{-0.5cm}
\end{figure}

Unfortunately, the zero-shot performance of the manually created templates from LAMA is low. For example, the BERT-large model only has around 30\% accuracy on the T-REx dataset. In our preliminary examinations, we found that in many error cases, \textit{the LM predicts the wrong type of objects}. We illustrate this in Figure \ref{fig:fewshot_motivation}. 

Motivated by this observation, in this work we explore few-shot knowledge probing, where only a small number (e.g., 10 or 20) of example triples are available to tune the prompts or model for better performance. This setting is attractive because: (1) Intuitively, a few examples are usually enough for humans to infer the precise relation type of interest; (2) Few-shot examples enable us to probe for new or rare relation types. 

In our experiments, we conduct a comprehensive comparison of different approaches in the context of few-shot knowledge probing. We briefly summarize our contributions as follows: (1) We create a new knowledge probing dataset named TREx-2p, which contains more challenging 2-hop relations. (2) For both 1-hop and 2-hop relations, few-shot examples strongly boost the knowledge probing performance for a pretrained LM. In particular, we find that a simple-yet-effective approach of finetuning the bias vectors in the model outperforms existing prompt-engineering methods. 


\vspace{-0.1cm}
\section{Few-shot Knowledge Probing}
\label{sec:method}
\vspace{-0.1cm}

We begin by establishing notations. We denote the parameters of a pretrained masked language model as $\theta$, and the vocabulary as $V$. For each relation type $r$, the probing dataset has a set of knowledge-base triples $\mathcal{D}^r = \{\langle x,r,y \rangle\}$, where $x$ and $y$ refers to the subject and object, respectively. Since we are considering a few-shot setting, each $\mathcal{D}^r$ is split to $\mathcal{D}_\text{fewshot}^r$ and $\mathcal{D}_\text{test}^r$, where $\mathcal{D}_\text{fewshot}^r$ only contain a small number (e.g., 10 or 20) of triples. Most of our approaches involve hyper-parameter tuning, in which case we further split $\mathcal{D}_\text{fewshot}^r$ into $\mathcal{D}_\text{train}^r$ and $\mathcal{D}_\text{dev}^r$. $\mathcal{D}_\text{dev}^r$ can also be used to prevent over-fitting (via early stopping).

The task of LM knowledge probing \citep{petroni2019knowlanguage} is to query a pretrained LM for $y$, by feeding it information about $x$ and $r$. To do so, a \textit{converter} function $f$ (to be described below) will be used to convert $x$ and $r$ into a query sentence with exactly one mask token in it, which is then fed into the LM. We denote the model's output distribution for the masked token as $P_\text{LM}(\,\cdot\ | f(x,r))$, and the performance is reflected by the rank of $y$ in that distribution. 

Next, we review available template options, and describe approaches which utilize the available few-shot training and development data to improve the performance of probing. Concrete examples are shown in Table \ref{tab:converter_example}.

\begin{table*}[]
\small
\centering
\def\arraystretch{1.1}
\addtolength{\tabcolsep}{-5.2pt}
\begin{tabular}{c|c}
\hline
$f$                              & $f(x,r)$                                                                                                    \\ \hline
manT/mineT       & Andrea Alciato was born in \texttt{<mask>} . / Andrea Alciato lived in \texttt{<mask>} .                                \\ \hline
defT         & Andrea Alciato => \texttt{<mask>} .                                                                         \\ \hline
manT+ctx & Joan Dickson was born in Edinburgh. Charles Helou was born in Beirut. Andrea Alciato was born in \texttt{<mask>}. \\ \hline
optiPrompts       & Andrea Alciato \texttt{<V0>} \texttt{<V1>} \texttt{<V2>} \texttt{<V3>} \texttt{<V4>} \texttt{<mask>}    \\ \hline
optiP+manT & Andrea Alciato \texttt{<V0>}:=was \texttt{<V1>}:=born \texttt{<V2>}:=in \texttt{<mask>} \texttt{<V3>}:=. \\ \hline 
\end{tabular}
\vspace{-0.2cm}
\caption{Examples of how different types of converters $f$ form a input for the masked language model. The relation $r$ is ``place of birth'', and the $x$ being queried is Andrea Alciato. The few-shot examples $\mathcal{D}_\text{train}^r$ consists of $\langle \text{Joan Dickson, Edinburgh}\rangle$ and $\langle \text{Charles Helou, Beirut}\rangle$, which can be used for in-context learning.}
\vspace{-0.25cm}
\label{tab:converter_example}
\end{table*}

\vspace{-0.1cm}
\paragraph{Template Options} In \citet{petroni2019knowlanguage}, the converter function (denoted by $f_\text{manT}$) is implemented via manually created templates, that are hand-crafted for each relation type. In \citet{jiang-etal-2020-know}, mining or paraphrasing-based methods are used to automatically find alternative templates for a given relation. They released the generated templates by the name LPAQA (LM Prompt And Query Archive), which are in the same format as in \citet{petroni2019knowlanguage}. For each relation $r$, we select the best-performing template by comparing the performance of each template on $\mathcal{D}_\text{fewshot}^r$, and use it for the convert function denoted by $f_\text{mineT}$. 

Both manT and mineT require human labor or external resources, therefore in the few-shot context setting we consider, it is reasonable to question whether such manual work is necessary. To explore this question, we follow \citet{NEURIPS2020_1457c0d6} and create a default template of ``[X]~=>~[Y]'', which can be applied for any relation type. We denote it by defT, and it can be used in the in-context learning approach described next.

\vspace{-0.1cm}
\paragraph{In-context Learning} As shown by \citep{NEURIPS2020_1457c0d6}, pretrained LMs are able to learn from the examples included in the input. To implement this approach, we concatenate converted triples from $\mathcal{D}_\text{train}^r$ to be a long prefix, and prepend it to our queries. We denote it by $f_\text{*+ctx}$, where $*$ is a placeholder for the template option (e.g., manT). In our experiments we find that the order of the prefix examples will affect the performance. Therefore for each relation type, we tune the ordering as a hyper-parameter via $\mathcal{D}_\text{dev}^r$.


\vspace{-0.1cm}
\paragraph{Optimized Prompts}  It is attractive to think of approaches which can automatically design prompts, minimizing human effort. AutoPrompt \citep{shin-etal-2020-autoprompt} and BERTese \citep{adi-2021-bertese} use gradient-based search to automatically find templates, in the form of discrete tokens, that maximize the model's performance on a training set. Very recently, OptiPrompt \citep{DBLP:journals/corr/abs-2104-05240} generalizes to continuous vectors, and achieves better performance than AutoPrompt. In OptiPrompt, five relation vectors are put between the subject and the mask token, before being fed into the model.\footnote{We have also tried with 8 or 10 relation vectors, but only observe very little improvements.} These relation vectors are trained to minimize the cross-entropy  loss for the object, with stochastic gradient descent (SGD):
\begin{equation}
    \mathcal{L}^r = - \frac{1}{|\mathcal{D}_\text{train}^r|}\sum_{\langle x,y\rangle \in \mathcal{D}_\text{train}^r} \log P_\text{LM}(y|f_\text{optiP}(x,r)).
\end{equation}
By default, we initialize the relation vectors to be the mean of the input embeddings of the first 10,000 most frequent tokens that are stored in the pretrained LM. We could also align the relation vectors with the manual template, and initialize them to be the embedding of the corresponding token in the template (denoted by optiP+manT). 

These studies utilize a considerable number of example triples (around 1000 samples per relation type) to train the prompts, and their performance under a few-shot setting is unknown.  

\vspace{-0.2cm}
\paragraph{Model Finetuning} All the approaches discussed above engineer the input while the pretrained LM is kept \textit{fixed}. Therefore, it is natural to consider  finetuning the model with the available templates or relation vectors as input. The major shortcoming is that we would need to store a copy of the entire model for each relation type \citep{brian2021scaleprompt,xiang2021prefixtuning}. 

\vspace{-0.2cm}
\paragraph{Model Bias Finetuning} To mitigate the storage issue, \citet{BenZaken2020BitFitSP} proposes to finetune only the bias vectors in the encoder. This approach is named \textit{BitFit}, and is shown to be very competitive on the GLUE benchmark \citep{wang-etal-2018-glue}. Further details and a storage cost comparison are given in Appendix \ref{sec:app_implement}. In our experiments, we test its performance under few-shot knowledge probing, and compare it with full-model finetuning. 


\vspace{-0.1cm}
\section{Datasets}
\label{sec:dataset}
\vspace{-0.1cm}

Following \citet{DBLP:journals/corr/abs-2104-05240} and \citet{shin-etal-2020-autoprompt}, we use the T-REx \citep{elsahar-etal-2018-rex} dataset, which is included in the LAMA benchmark \citep{petroni2019knowlanguage}. It contains 41 Wikidata relation types, and each relation type has up to 1000 triples. We will refer to it as TREx-1p as it focuses on 1-hop relations.

In addition to memory of 1-hop relations, humans also possess the capability of multi-hop reasoning \citep{yang-etal-2018-hotpotqa,xiong2017deeppath}. For example, given two known facts of "[X] works for [Y]." and "[Y] produces [Z].", there is clearly a 2-hop link between X and Z (e.g., X being Steve Jobs, Y being Apple, and Z being iPhone). To probe whether the pretrained LM also possesses this kind of ``indirect'' knowledge, we create a 2-hop variant of the T-REx dataset, named \textbf{TREx-2p}.\footnote{We will release the data and code used in this work in the public version of this manuscript.} We manually examine the 2-hop link existing in the knowledge graph of TREx-1p, and select eight 2-hop relation types that make sense to humans. 

As in LAMA, we manually create natural-language templates for relations in TREx-2p. We show them in Table \ref{tab:trex2p_temp} (Appendix \ref{sec:app_aux}). To encode the 2-hop relations, these templates are syntactically more complicated (e.g., ``[X] works for a company that developed [Y] .''). Therefore, we expect the zero-shot probing performance of TREx-2p with manual templates to be low.

\vspace{-0.1cm}
\section{Experiments}
\label{sec:exp}
\vspace{-0.1cm}

Our experiments focus on the Roberta-large model \citep{yinhan19roberta}, a 24-layer transformer LM with a hidden-dimension of 1024. Our code is based on HuggingFace \citep{wolf-etal-2020-transformers} and the released code from LAMA. The few-shot development set ($\mathcal{D}_\text{dev}$) are used for hyper-parameter tuning. We find that finetuning with the few-shot training examples are very prone to over-fitting. Therefore, during SGD finetuning we do early stopping by monitoring the loss on the development set every 10 iterations. More details are in Appendix \ref{sec:app_implement}.

In addition to accuracy (Precision@1), we also report mean reciprocal rank (MRR), to account for cases with multiple valid targets. Following earlier work \citep{petroni2019knowlanguage}, we report macro-averaged numbers across different relation types.

The few-shot examples are randomly selected from the dataset for each relation type, and the rest of the samples are used for evaluation. We compare the performance of different approaches under settings where 10/20/40 example triples are available. Out of the available examples, 5/10/10 samples are taken out as a development set, leaving the rest for training. The same training/development sets are used across different approaches. 

\begin{table*}[t]
\small
\centering
\def\arraystretch{1.1}
\addtolength{\tabcolsep}{-2.5pt}
\begin{tabular}{|c|cccc|ccc|cc|ccc|}
\hline
\textbf{Accuracy(\%)} & \multicolumn{4}{c|}{\textbf{Prompt Engineering}}                                                                          & \multicolumn{3}{c|}{\textbf{In-context Learning}} & \multicolumn{2}{c|}{\textbf{Model FT}} & \multicolumn{3}{c|}{\textbf{BitFit}} \\ \hline
\textbf{TREx-1p}      & manT                                                                                         & mineT & optiP & optiP+manT & manT           & defT           & mineT           & manT              & defT                       & manT          & defT & optiP+manT    \\ \hline
5T+5D                 & \multicolumn{1}{c|}{\multirow{3}{*}{\begin{tabular}[c]{@{}c@{}}0-shot:\\ 25.8\end{tabular}}} & 34.9  & 40.0  & 49.4       & 49.0           & 47.3           & 48.9            & 49.1              & 44.8                       & 49.2          & 45.4 & \textbf{49.8} \\ \cline{1-1} \cline{3-13} 
10T+10D               & \multicolumn{1}{c|}{}                                                                        & 36.3  & 47.9  & 49.7       & 50.3           & 51.1           & 51.6            & 51.3              & 49.4                       & \textbf{52.4} & 48.9 & 52.1          \\ \cline{1-1} \cline{3-13} 
30T+10D               & \multicolumn{1}{c|}{}                                                                        & 37.0  & 52.3  & 52.5       & 50.0           & 52.1           & 51.0            & 54.1              & 53.2                       & \textbf{54.5} & 53.3 & 54.0             \\ \hline \hline
\textbf{TREx-2p}      & manT                                                                                         & /     & optiP & optiP+manT & manT           & defT           & /               & manT              & defT                       & manT          & defT & optiP         \\ \hline
5T+5D                 & \multicolumn{1}{c|}{\multirow{3}{*}{\begin{tabular}[c]{@{}c@{}}0-shot:\\ 14.4\end{tabular}}} & /     & 43.2  & 41.3         & 47.5           & 45.0           & /               & 45.6              & \textbf{48.1}              & 44.6          & 46.9          & 48.0          \\ \cline{1-1} \cline{3-13} 
10T+10D               & \multicolumn{1}{c|}{}                                                                        & /     & 50.1  & 46.7         & 44.0           & 44.0           & /               & 50.1              & 48.9                       & 51.4          & \textbf{51.5} & 50.1          \\ \cline{1-1} \cline{3-13} 
30T+10D               & \multicolumn{1}{c|}{}                                                                        & /     & 51.8  & 52.0        & 53.0           & 50.3           & /               & 53.5              & 54.2                       & 53.6          & 53.5          & \textbf{55.7} \\ \hline
\end{tabular}
\vspace{-0.2cm}
\caption{The accuracy performance of different approaches for the TREx-1/2p datasets. ``5T+5D'' means that 5 examples are used for training and 5 examples are used as a development set. Some combination of approaches (such as model finetuning with OptiPrompt) are deferred to Appendix \ref{sec:app_aux} due to lack of space.}
\vspace{-0.3cm}
\label{tab:trex_main_acc}
\end{table*}

The accuracy results are shown in Table \ref{tab:trex_main_acc}. Observations from results measured by MRR are highly similar, and we defer them to Table \ref{tab:trex_main_mrr} (Appendix \ref{sec:app_aux}) to save space. In general, we observe that for both 1-hop and 2-hop relations, large gains can be achieved with as few as 10 available examples in comparison to the zero-shot performance. 


For prompt engineering, OptiPrompt greatly outperforms manual or LPAQA (mineT) templates, which agrees with the non-few-shot results in \citet{DBLP:journals/corr/abs-2104-05240}. This confirms the advantage of a continuous prompt as opposed to discrete tokens. Next, in-context learning is  competitive in the 10/20-shot setting. However, its performance saturates quickly, and is outperformed by OptiPrompt in the 40-shot setting. 

Direct finetuning a large model with only a few training examples is usually considered difficult due to over-fitting. Interestingly, we find that early stopping with the tiny development set can effectively regularize the training, and model fintuning gives better accuracy than OptiPrompt in most cases. More excitingly, BitFit, which only tunes the bias parameters, achieves similar or even better accuracy than full-model finetuning. In some cases, BitFit can benefit from OptiPrompt as input for the extra flexibility. Lastly, we observe that manual templates perform better than the default template for OptiPrompt, model finetuning and BitFit, showing a complementary effect to the few-shot examples.  

Will more examples give better performance? In Appendix \ref{sec:app_aux}, we show that the performance saturates at around 200 examples with an accuracy of 57.5\%.

\begin{figure}
    \centering
    \includegraphics[width=0.85\linewidth]{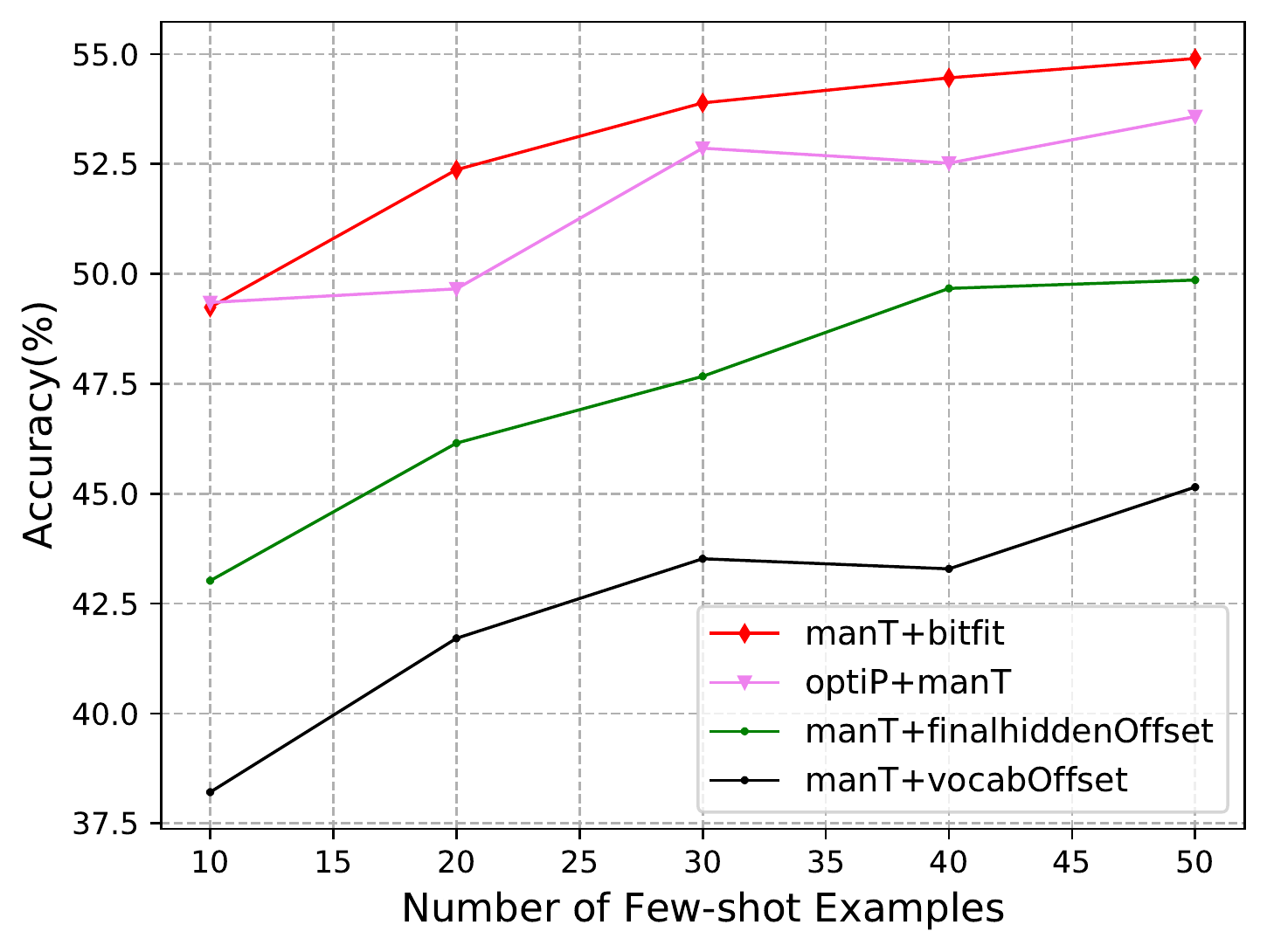}
    \vspace{-0.2cm}
    \caption{(TREx-1p) Only finetuning the bias parameter in the output layer or the final hidden layer gives worse performance than BitFit or OptiPrompt.}
    \label{fig:fitbit_control}
    \vspace{-0.3cm}
\end{figure}

For TREx-2p, the general observations are similar to TREx-1p. We mention two differences: (1) The zero-shot performance of manual templates for TREx-2p is poor (only 14.4\%), which is expected as 2-hop templates are syntactically more complicated; (2) Possibly due to the same reason, OptiPrompt does not benefit from manual-template initialization for TREx-2p.

Finally, we introduce two control baselines for BitFit: (1) Only the length-$|V|$ final bias vector in the output layer is finetuned; (2) Only the bias vector in the final hidden layer is finetuned. Results are shown in Figure \ref{fig:fitbit_control}. We observe that both control baselines are outperformed by BitFit and OptiPrompt by a large margin. This shows that the performance gain is not only from simply biasing the model to a certain group of output tokens, and the inner representations are also changed to better expose the stored knowledge.

\vspace{-0.1cm}
\section{Related Works}
\vspace{-0.1cm}


How to effectively adapt a pretrained LM to a specific task in a few-shot setting has been an important topic in recent NLP research \citep{zhang2021revisiting}. The idea of in-context learning is popularized by GPT-3 \citep{NEURIPS2020_1457c0d6}, which shows that a fixed pretrained LM can be \textit{primed} to conduct different tasks via in-context examples. Recently, \citet{tony2021califewshot} points out some caveats about in-context learning, and how to better calibrate it. 

Closely related to template-based knowledge probing, \citet{schick-schutze-2021-exploiting} proposes Pattern Exploiting Training (PET) for NLU tasks, where inputs are converted into cloze-style questions (e.g., ``Awful pizza! It was \texttt{<mask>}.''), and gradient-based optimization is conducted. PET and its variants iPET and ADAPET \citep{timo2020smallfewshot,derek2021adapet} are shown to be more effective than vanilla in-context learning. Also along this line of work, \citet{tianyu2020demonstratefewshot} propose an automatic framework of prompt generation and demonstration selection. 

Last but not least, \citet{xiang2021prefixtuning}, followed by \citet{brian2021scaleprompt}, propose \textit{prefix tuning}, where continuous task-specific input vectors are tuned while the model is kept fixed. It is very similar to the OptiPrompt approach considered in this work.



\bibliography{emnlp2021}
\bibliographystyle{acl_natbib}

\clearpage
\appendix

\section*{Appendices}

\section{Implementation Details}
\label{sec:app_implement}

For model or relation vector finetuning, we use the AdamW optimizer \citep{loshchilov2018decoupled,zhang2021revisiting}. Since in most of our few-shot experiments, the training data only consists of a small number (e.g., 10 or 20) of samples, we directly do full-batch training. We tune the learning-rates in a log-scale using $\mathcal{D}_\text{dev}^r$. Typically, we find that a small learning rate (e.g., 1e-06) works well for full-model finetuning, while a relatively large learning rate (e.g., 0.01) works well for OptiPrompt or BitFit. For in-context learning, we try 20 different random orders (of the examples in the context) for each relation type, and use the ordering which gives best performance on $\mathcal{D}_\text{dev}^r$.

For the implementation of BitFit, we follow the BitFit-$\partial$ variant in \citet{BenZaken2020BitFitSP}, where around half of the bias vectors are tuned. To be specific, for each transformer layer, the bias for the attention query (of length 1024), and the bias for the intermediate layer (of length 4096) of the transformer block, are tuned. We summarize and compare the storage cost of different approaches in Table \ref{tab:param_number}.

Finally, we mention two differences between our implementation and the code from the LAMA benchmark: (1) The original code uses a \textit{common vocab}, which is a intersection of the vocabularies from various pretrained LMs. In this work since we focus on the Roberta-large model, we just use the whole Roberta vocab. If we switch to the common vocab, the zero-short accuracy on the TREx-1p dataset will be improved from 25.8\% to 31.9\%. (2) In the original code, inside each relation, if a subject has multiple valid objects, they are still treated as separate triples. As a consequence, for the accuracy metric, it is impossible for the LM get all triples right because only the top-1 prediction is considered. In our implementation, we merge them into one test case with multiple valid targets, and report both accuracy and MRR.

\section{Auxiliary Results}
\label{sec:app_aux}

Templates created for TREx-2p are shown in Table \ref{tab:trex2p_temp}. These 2-hop relations are manually selected from the knowledge graph of TREx-1p.

In Table \ref{tab:trex_main_mrr}, MRR performance of different approaches are shown. The observations are similar to the accuracy results (Table \ref{tab:trex_main_acc}). 

\begin{figure}
    \centering
    \includegraphics[width=0.85\linewidth]{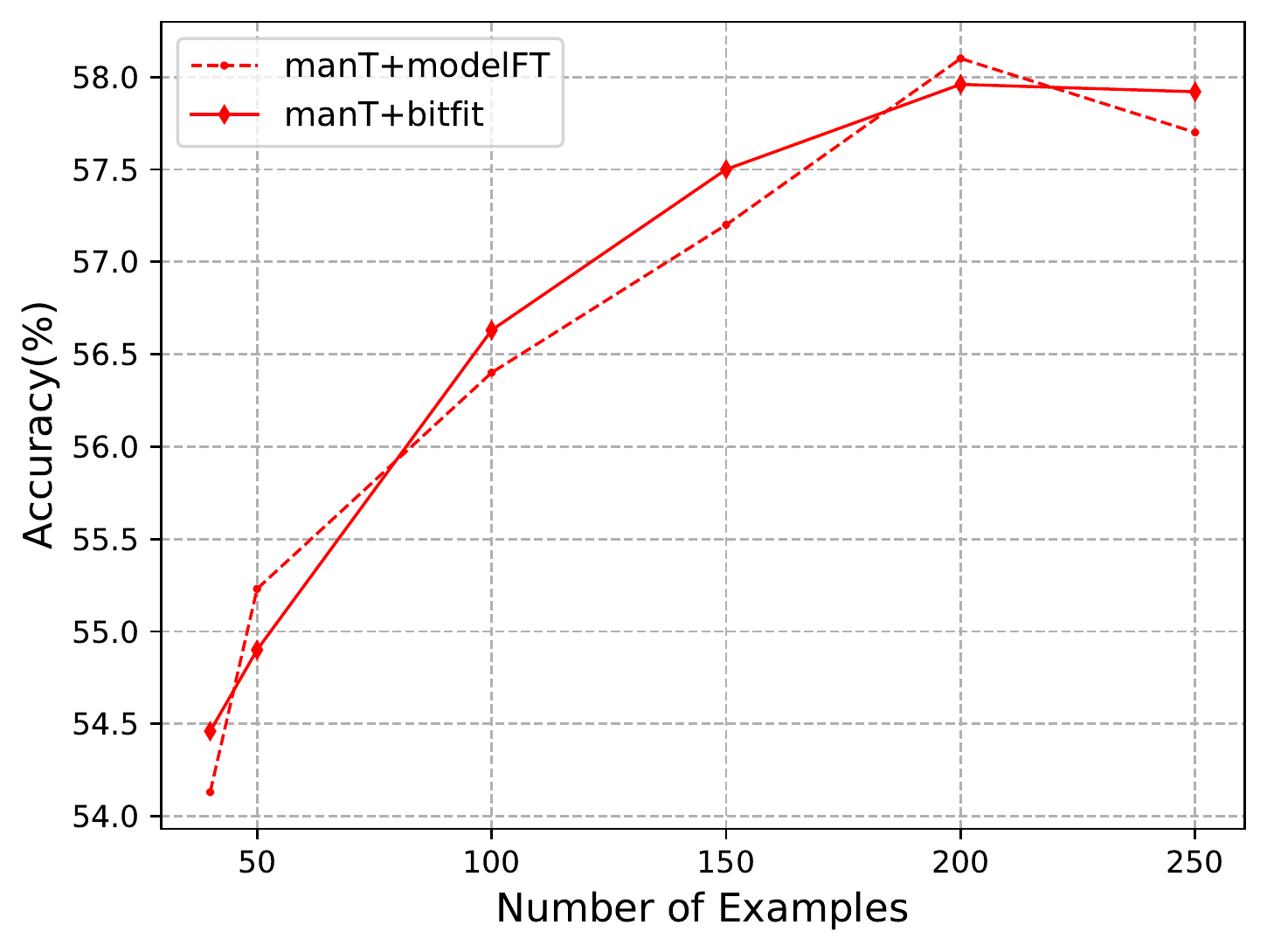}
    \caption{(TREx-1p) Accuracy results with more available samples for each relation, the performance saturates at around 200 samples.}
    \label{fig:trex_moresample}
\end{figure}

In Figure \ref{fig:trex_moresample}, accuracy results with more available samples are shown. We observe that the performance saturates at around 200 samples at around 57.5\%. When the number of available samples is larger than or equal to 100, we use 20 samples for development.

In Table \ref{tab:trex_combine}, a complete set of results of model fintuning/BitFit for the TREx-1/2p datasets are shown. OptiPrompt and the model (or the bias vectors in the model) can be jointly trained for the extra flexibility in the input. In some cases the performance is improved, but the gain is not large.

\begin{table}[]
\small
\centering
\begin{tabular}{|c|c|}
\hline
\textbf{Approach} & \textbf{Param. Number} \\
\hline
In-context Leanring & 0 \\
OptiPrompt & 5 $\times$ 1024 \\
BitFit & 24 $\times$ 5120 \\
Model Finetuning & ~355M \\
\hline
FinalHiddenOffset & 1024 \\
VocabOffset & |V| = 50325  \\
\hline
\end{tabular}
\caption{The number of extra parameters to be saved for each relation type.}
\label{tab:param_number}
\end{table}

\begin{table*}[]
\small
\centering
\begin{tabular}{l}
\hline
P159 The headquarter of [X] (Virtue Party) is in [Y] (Ankara) . | P1376 [X] (Ankara) is the capital of [Y] (Turkey) . \\
\textbf{The headquarter of [X] (Virtue Party) is in the country of [Y] (Turkey) .} \\
\hline
P108 [X] (Steve Jobs) works for [Y] (Apple) . | P178 [X] (macOS) is developed by [Y] (Apple) .\\
\textbf{[X] (Steve Jobs) works for a company that developed [Y] (macOS) .} \\
\hline
P178 [X] (macOS) is developed by [Y] (Apple) . | P178 [X] (MessagePad) is developed by [Y] (Apple) . \\
\textbf{[X] (macOS) and [Y] (MessagePad) are developed by the same company .} \\ \hline
P31 [X] (Wick Airport) is a [Y] (airport) . | P361 [X] (runway) is part of [Y] (airport) . \\
\textbf{One component of [X] (Wick Airport) is [Y] (runway) .} \\ \hline

P361 [X] (geometry) is part of [Y] (mathematics) . | P361 [X] (arithmetic) is part of [Y] (mathematics) . \\
\textbf{[X] (geometry) and [Y] (arithmetic) are part of the same thing .} \\ \hline
P361 [X] (whey) is part of [Y] (milk) . | P527 [X] (yogurt) consists of [Y] (milk) . \\
\textbf{[X] (whey) is a low-level part of [Y] (yogurt) .} \\ \hline
P527 [X] (gelato) consists of [Y] (milk) . | P527 [X] (yogurt) consists of [Y] (milk) . \\
\textbf{[X] (gelato) and [Y] (yogurt) share at least one element .} \\ \hline
P37 The official language of [X] (Scotland) is [Y] (English) . | P19 [X] (Paul Mounsey) was born in [Y] (Scotland) . \\
\textbf{The official language of the country where [X] (Paul Mounsey) was born is [Y] (English) .} \\ \hline
\end{tabular}
\caption{Examples of the TREx-2p dataset, and the manual templates we created. The relation ids from the origin T-REx dataset are also shown.}
\label{tab:trex2p_temp}
\end{table*}

\begin{table*}[t]
\small
\centering
\def\arraystretch{1.1}
\addtolength{\tabcolsep}{-2.5pt}
\begin{tabular}{|c|cccc|ccc|cc|ccc|}
\hline
\textbf{MRR}     & \multicolumn{4}{c|}{\textbf{Prompt Engineering}}                                                                          & \multicolumn{3}{c|}{\textbf{In-context Learning}} & \multicolumn{2}{c|}{\textbf{Model Finetuning}} & \multicolumn{3}{c|}{\textbf{BitFit}} \\ \hline
\textbf{TREx-1p} & manT                                                                                         & mineT & optiP & optiP+manT & manT           & defT           & mineT           & manT                       & defT              & manT          & defT & optiP+manT    \\ \hline
5T+5D            & \multicolumn{1}{c|}{\multirow{3}{*}{\begin{tabular}[c]{@{}c@{}}0-shot:\\ .340\end{tabular}}} & .436  & .487  & .572       & .568           & .559           & .572            & .576                       & .538              & .577          & .538 & \textbf{.580} \\ \cline{1-1} \cline{3-13} 
10T+10D          & \multicolumn{1}{c|}{}                                                                        & .450  & .559  & .577       & .583           & .596           & .596            & .594                       & .580              & \textbf{.600} & .574 & \textbf{.600} \\ \cline{1-1} \cline{3-13} 
30T+10D          & \multicolumn{1}{c|}{}                                                                        & .458  & .603  & .608       & .583           & .603           & .595            & \textbf{.626}              & .616              & .625          & .617 & .625          \\ \hline \hline
\textbf{TREx-2p} & manT                                                                                         & /     & optiP & optiP+manT & manT           & defT           & /               & manT                       & defT              & manT          & defT & optiP         \\ \hline
5T+5D            & \multicolumn{1}{c|}{\multirow{3}{*}{\begin{tabular}[c]{@{}c@{}}0-shot:\\ .166\end{tabular}}} & /     & 35.8  & 35.0       & \textbf{44.9}        & 42.7        & /            & 39.3                       & 40.0              & 38.0          & 38.9 & 43.1          \\ \cline{1-1} \cline{3-13} 
10T+10D          & \multicolumn{1}{c|}{}                                                                        & /     & 42.6  & 42.3       & 41.3                 & 41.2        & /            & 44.6                       & 43.4              & \textbf{46.5} & 44.3 & 43.3          \\ \cline{1-1} \cline{3-13} 
30T+10D          & \multicolumn{1}{c|}{}                                                                        & /     & 44.4  & 43.7       & 44.2                 & 44.5        & /            & 45.8                       & 45.9              & 46.3          & 44.8 & \textbf{46.6} \\ \hline
\end{tabular}
\caption{The MRR performance of different approaches for the TREx-1/2p datasets. The leading zeros are omitted. The observations are similar to the accuracy results. \label{tab:trex_main_mrr}}
\end{table*}

\begin{table*}[t]
\small
\centering
\def\arraystretch{1.1}
\addtolength{\tabcolsep}{-2.5pt}

\begin{tabular}{|c|ccccc|ccccc|}
\hline
\textbf{Accuracy(\%)} & \multicolumn{5}{c|}{\textbf{Model Finetuning}}    & \multicolumn{5}{c|}{\textbf{Bitfit}}                                 \\ \hline
\textbf{TREx-1p}      & manT & defT          & mineT & optiT & optiT+manT & manT          & defT & mineT         & optiT         & optiP+manT    \\ \hline
5T+5D                 & 49.1 & 44.8          & 48.7  & 42.6  & 49.0       & 49.2          & 45.4 & 48.8          & 44.3          & \textbf{49.8} \\ \hline
10T+10D               & 51.3 & 49.4          & 51.1  & 49.1  & 51.2       & \textbf{52.4} & 48.9 & 51.1          & 47.7          & 52.2          \\ \hline
30T+10D               & 54.1 & 53.2          & 54.1  & 53.3  & 54.2       & \textbf{54.5} & 53.3 & \textbf{54.5} & 53.1          & 54.0          \\ \hline
\textbf{TREx-2p}      & manT & defT          & mineT & optiT & optiT+manT & manT          & defT & mineT         & optiT         & optiP+manT    \\ \hline
5T+5D                 & 45.6 & \textbf{48.1} & /     & 44.3  & 46.4       & 44.6          & 46.9 & /             & 48.0          & 45.3          \\ \hline
10T+10D               & 50.1 & 48.9          & /     & 50.0  & 48.8       & \textbf{51.4} & 51.5 & /             & 50.1          & 50.6          \\ \hline
30T+10D               & 53.5 & 54.2          & /     & 52.5  & 53.6       & 53.6          & 53.5 & /             & \textbf{55.7} & 53.6          \\ \hline
\textbf{MRR}          & \multicolumn{5}{c|}{\textbf{Model Finetuning}}    & \multicolumn{5}{c|}{\textbf{Bitfit}}                                 \\ \hline
\textbf{TREx-1p}      & manT & defT          & mineT & optiT & optiT+manT & manT          & defT & mineT         & optiT         & optiP+manT    \\ \hline
5T+5D                 & .576 & .538          & .569  & .514  & .575       & .577          & .538 & .570          & .532          & \textbf{.580} \\ \hline
10T+10D               & .594 & .580          & .590  & .570  & .594       & \textbf{.600} & .574 & .590          & .562          & \textbf{.600} \\ \hline
30T+10D               & .626 & .616          & .623  & .613  & \textbf{.627}       & .625          & .617 & .626 & .612          & .625          \\ \hline
\textbf{TREx-2p}      & manT & defT          & mineT & optiT & optiT+manT & manT          & defT & mineT         & optiT         & optiP+manT    \\ \hline
5T+5D                 & 39.3 & 40.0          & /     & 37.3  & 39.6       & 38.0          & 38.9 & /             & \textbf{43.1} & 39.3          \\ \hline
10T+10D               & 44.6 & 43.4          & /     & 42.8  & 43.2       & \textbf{46.5} & 44.3 & /             & 43.3          & 46.0          \\ \hline
30T+10D               & 45.8 & 45.9          & /     & 44.7  & 45.6       & 46.3          & 44.8 & /             & \textbf{46.6} & 45.9          \\ \hline
\end{tabular}

\vspace{-0.2cm}
\caption{A complete set of results of model fintuning and BitFit for the TREx-1/2p datasets.}
\vspace{-0.2cm}
\label{tab:trex_combine}
\end{table*}

\end{document}